\documentclass{article}

\usepackage{arxiv}

\usepackage[utf8]{inputenc} 
\usepackage[T1]{fontenc}    
\usepackage{hyperref}       
\usepackage{url}            
\usepackage{booktabs}       
\usepackage{amsfonts}       
\usepackage{nicefrac}       
\usepackage{microtype}      
\usepackage{lipsum}
\usepackage{graphicx}
\usepackage{amsmath}
\usepackage[pdf]{graphviz}
\usepackage{tikz}
\usepackage{subcaption} 
\usepackage{hyperref}

\graphicspath{ {./images/} }

\title{Lightweight Metadata-Aware Mixture-of-Experts Masked Autoencoder for Earth Observation}

\author{
 Mohanad Albughdadi \\
  European Centre for Medium-Range Weather Forecasts\\
  Bonn, Germany 53175 \\
  \texttt{mohanad.albughdadi@ecmwf.int} \\
}

\begin{document}
\maketitle
\begin{abstract}
Recent advances in Earth Observation have focused on large-scale foundation models. However, these models are computationally expensive, limiting their accessibility and reuse for downstream tasks. In this work, we investigate compact architectures as a practical pathway toward smaller general-purpose EO models. We propose a Metadata-aware Mixture-of-Experts Masked Autoencoder (MoE-MAE) with only ~2.5M parameters. The model combines sparse expert routing with geo-temporal conditioning, incorporating imagery alongside latitude/longitude and seasonal–daily cyclic encodings. We pretrain the MoE-MAE on the BigEarthNet-Landsat dataset and evaluate embeddings from its frozen encoder using linear probes. Despite its small size, the model competes with much larger architectures, demonstrating that metadata-aware pretraining improves transfer and label efficiency. To further assess generalization, we evaluate on the EuroSAT-Landsat dataset, which lacks explicit metadata, and still observe competitive performance compared to models with hundreds of millions of parameters. These results suggest that compact, metadata-aware MoE-MAEs are an efficient and scalable step toward future EO foundation models. Model's code and weights are available on this repository \url{https://github.com/AlbughdadiM/geo-moe-mae}.   
\end{abstract}


\section{Introduction}
\label{sec:intro}
Supervised learning has driven advances in machine learning~(ML) applications for Earth Observation~(EO), \textit{e.g.}, scene classification, land cover / land use classification, and object or change detection. However, training such algorithms requires large amounts of carefully prepared labeled data, which is challenging to obtain in EO due to geographical, temporal, and cross-sensor variability.  

Large-scale self-supervised pretraining has emerged as a promising alternative. Models such as SatMAE~\cite{cong2023satmaepretrainingtransformerstemporal}, Prithvi~\cite{jakubik2023foundationmodelsgeneralistgeospatial}, TerraMind~\cite{jakubik2025terramindlargescalegenerativemultimodality} and Clay~\cite{clay2024} have shown that ML algorithms can learn transferable embeddings that generalize across diverse downstream tasks. While advanced tasks may still require finetuning, these benefits come at the cost of very large model sizes (hundreds of millions of parameters) and substantial pretraining budgets. Such requirements limit reproducibility, accessibility, and adoption in resource-constrained settings.  

A natural question is whether smaller architectures, trained with the right objectives and inductive biases, can capture broadly reusable EO representations while remaining lightweight and efficient. Compact models would democratize access to pretrained EO representations, lower the barrier for downstream applications, and enable scalable deployment in real-world scenarios.  

This work explores this question by designing a metadata-aware Mixture-of-Experts Masked Autoencoder~(MoE-MAE) with only $\sim$2.5M parameters and its encoder with only $\sim$2.3M parameters. Our model combines two ideas:
\begin{enumerate}
    \item \textbf{Sparse expert routing}: tokens are processed by only a small subset of experts, enabling specialization while keeping active compute low.
    \item \textbf{Geo-temporal conditioning}: image features are fused with latitude/longitude and cyclical time encodings (seasonal and daily), allowing the model to exploit spatio-temporal regularities inherent in EO data.
\end{enumerate}

The MoE-MAE is pretrained on the BigEarthNet-Landsat dataset using a self-supervised objective that combines masked patch reconstruction with auxiliary losses on unmasked patches and a balancing loss for mixture-of-experts routing. To evaluate the learned representations, we train linear probes on frozen embeddings. Despite its compact size, our model achieves the following.
\begin{enumerate}
    \item \textbf{On BigEarthNet}: competitive multi-label classification accuracy compared to much larger dense models.
    \item \textbf{On EuroSAT}: good performance despite the dataset lacking explicit metadata, demonstrating that the learned embeddings generalize beyond pretraining conditions.
\end{enumerate}

These results suggest that general EO models could be achieved with compact, metadata-aware MoE-MAEs, challenging the notion that only massive parameter counts lead to strong transfer.

\section{Related Work}
\label{sec:related_work}

Self-supervised learning (SSL) is a pretraining paradigm that allows models to learn from large unlabeled datasets. This has become central in EO, and has been adopted in multiple works, \textit{e.g.}, SatMAE~\cite{cong2023satmaepretrainingtransformerstemporal}, Prithvi~\cite{jakubik2023foundationmodelsgeneralistgeospatial}, TerraMind~\cite{jakubik2025terramindlargescalegenerativemultimodality}, Clay~\cite{clay2024}, and SSL4EO~\cite{wang2023ssl4eos12largescalemultimodalmultitemporal,stewart2023ssl4eoldatasetsfoundationmodels}. Most of these models adopt transformer-based architectures and are pretrained on global multi-sensor archives, serving as general-purpose backbones for EO tasks. They transfer successfully to land cover mapping, change detection, and semantic segmentation. However, their benefits come with substantial computational costs, as they contain tens or hundreds of millions of parameters and require access to large-scale compute clusters. This limits accessibility, reproducibility, and adoption in resource-constrained environments, and highlights the need for compact models that remain transferable.  

\paragraph{Mixture-of-Experts.}  
Mixture-of-Experts (MoE) architectures increase model capacity by routing tokens to a small subset of experts, thus activating only part of the network per forward pass. This idea, introduced in NLP with Switch Transformers~\cite{fedus2022switchtransformersscalingtrillion}, has since been extended to vision tasks~\cite{riquelme2021scalingvisionsparsemixture}. MoE has also been combined with masked autoencoding in the recently proposed mLiT/mmLiT~\cite{tan2024lightweightvisiontransformer}, which introduced lightweight MoE-MAEs for small-scale RGB datasets. However, mLiT was not applied to Earth Observation, did not release code, and did not investigate metadata-aware learning.

\paragraph{Metadata-aware representation learning in EO.}  
Unlike natural images, EO data are inherently tied to space and time. Several works incorporate metadata such as scale~\cite{reed2023scalemaescaleawaremaskedautoencoder}, sensor parameters~\cite{prexl2024senpamaesensorparameteraware}, and geographic anchors,into MAE-based models. For instance, Clay~\cite{clay2024} integrate spatial, temporal, and sensor parameters to learn embeddings. These approaches demonstrate that metadata can improve representation quality, but they are built on dense transformers and do not exploit the efficiency benefits of MoE routing.

\paragraph{Our contribution.}  
To the best of our knowledge, this is the first work to extend a Mixture-of-Experts MAE to satellite imagery by incorporating geo-temporal metadata directly into pretraining. Building on mLiT/mmLiT~\cite{tan2024lightweightvisiontransformer}, we adapt the lightweight MoE-MAE design to EO and introduce spatio-temporal conditioning (latitude/longitude and seasonal–daily encodings). We further evaluate transfer on standard EO benchmarks (BigEarthNet, EuroSAT). By doing so, we present the first compact, metadata-aware MoE-MAE for EO, demonstrating that strong transferability can be achieved without massive parameter counts.

\section{Method}
\label{sec:method}
\begin{figure}[ht]
  \centering
  \includegraphics[width=\linewidth]{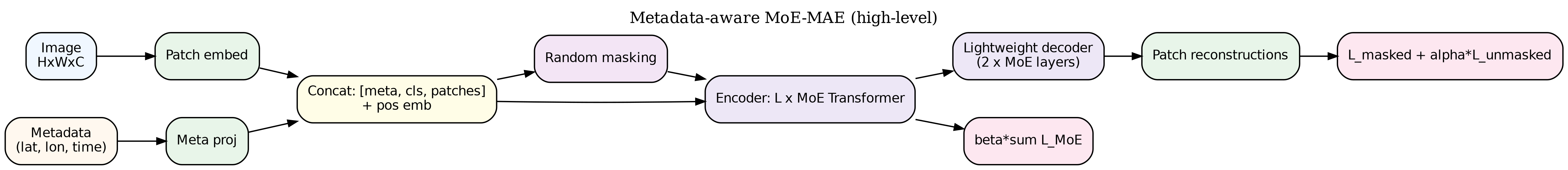}
  \caption{High-level architecture of the metadata-aware MoE-MAE.}
  \label{fig:moe_mae_arch}
\end{figure}

MAE is an effective pretraining strategy for vision transformers~\cite{he2021maskedautoencodersscalablevision}. It relies on masking a large portion of input patches and train the model to reconstruct them from the tokens of the unmasked patches (visible tokens). This forces the encoder to learn semantically meaningful representations that capture spatial and spectral structure. Given an input image $\mathbf{x} \in \mathbb{R}^{H \times W \times C}$, the image is divided into non-overlapping patches of size $p \times p$. These patches are linearly projected to obtain a sequence of embeddings. A portion of the patches is then randomly masked using a fixed ratio $r$. Only visible tokens are passed to the encoder $f_\theta(\cdot)$. Finally, a lightweight decoder $g_\phi(\cdot)$ receives both encoded visible tokens and masked token placeholders, and attempts to reconstruct the full input. The proposed model architecture is depicted in Fig.~\ref{fig:moe_mae_arch}.

\subsection{Encoder}
The encoder builds upon a compact Vision Transformer with Mixture-of-Experts (MoE) layers.  

\paragraph{Patch embedding.}  
An input image $\mathbf{X} \in \mathbb{R}^{H \times W \times C}$ is divided into non-overlapping patches of size $p \times p$, which are linearly projected into an embedding space of dimension $d$.  

\paragraph{Metadata tokens.}  
Each image is associated with geo-temporal metadata: latitude, longitude, week-of-year, and hour-of-day. These attributes are encoded as sinusoidal pairs $(\sin, \cos)$ to preserve cyclic structure, and then projected into the embedding dimension through linear layers. This produces four metadata tokens $\mathbf{t}_{\text{week}}, \;\mathbf{t}_{\text{hour}}, \;\mathbf{t}_{\text{lat}}, \;\mathbf{t}_{\text{lon}} \in \mathbb{R}^d$.
They are concatenated with the class token and patch embeddings at the encoder input:
\begin{equation}
\mathbf{z}_0 = [\mathbf{t}_{\text{week}}, \mathbf{t}_{\text{hour}}, \mathbf{t}_{\text{lat}}, \mathbf{t}_{\text{lon}}, \mathbf{t}_{\text{cls}}, \mathbf{t}_1, \dots, \mathbf{t}_N] + \mathbf{P},
\end{equation}
where $\mathbf{P}$ denotes positional embeddings.
\paragraph{Mixture-of-Experts transfomer encoder layer.}
Each encoder block follows a pre-norm residual architecture with two sublayers. Layer normalization is first applied to the inputs and the output is passed through Grouped Query Attention~(GQA), which reduces the dimensionality of keys and values by grouping attention heads to reduce costs while maintaining the multi-head attention mechanism. A residual connection then adds the attention output back to the input. Then another normalization is applied to the results before passing the output to MoE feed-forward network. An MoE layer consists of:
    \begin{enumerate}
    \item{\textit{SwiGLU experts:}}  
    The feed-forward network in each transformer block is replaced by an MoE module composed of multiple SwiGLU experts~\cite{shazeer2020gluvariantsimprovetransformer}. To reduce parameters, the projection matrices $\mathbf{V}$ and $\mathbf{W}_2$ can be shared across experts, following mLiT~\cite{tan2024lightweightvisiontransformer}. This design increases representational power while keeping the model compact.

    \item{\textit{Routing with NoisyTop-$k$:}}  
    A lightweight gating network assigns tokens to experts using NoisyTop-$k$ routing~\cite{shazeer2017outrageouslylargeneuralnetworks}. For each token embedding $\mathbf{z}$, the router produces noisy logits
    \begin{equation}
    \mathbf{H} = \mathbf{W}_g \mathbf{z} + \boldsymbol{\epsilon} \odot \text{softplus}(\mathbf{W}_{\text{noise}} \mathbf{z}),
    \end{equation}
    where $\boldsymbol{\epsilon} \sim \mathcal{N}(0,1)$. The top-$k$ logits in $\mathbf{H}$ are selected, and a softmax over them yields sparse routing weights. Each token is thus processed by only a small subset of experts, keeping active compute low while allowing specialization.
    \item{\textit{Load-balancing regularization:}}  
    To prevent collapse caused by imbalanced expert utilization due to naive MoE routing, two scalar penalties based on the coefficient of variation (CV) are used:
    \begin{equation}
    \mathcal{L}_{\text{imp}} = \lambda_{\text{imp}} \cdot \mathrm{CV}\!\Big(\sum_i G(x_i)\Big),
    \qquad
    \mathcal{L}_{\text{load}} = \lambda_{\text{load}} \cdot \mathrm{CV}\!\Big(\sum_i P(x_i)\Big),
    \end{equation}
    where $G(x_i)$ is the routing gate activations for token $i$ and $P(x_i)$ is the expected noisy routing probabilities. The total MoE regularization is:
    \begin{equation}
    \mathcal{L}_{\text{MoE}} = \mathcal{L}_{\text{imp}} + \mathcal{L}_{\text{load}}. 
    \end{equation}
    \end{enumerate}
\paragraph{Compact MoE design.}  
MoE layers are inserted throughout the depth of the encoder, with the number of experts gradually increasing across stages, following the staged expert configuration of~\cite{tan2024lightweightvisiontransformer}. This progressive scaling allows early layers to remain lightweight while deeper layers gain more capacity. During training, each MoE layer contributes a regularization term $\mathcal{L}_{\text{MoE}}^{(\ell)}$, and the total loss is accumulated across all $L$ encoder layers:
\begin{equation}
\mathcal{L}_{\text{MoE}} = \sum_{\ell=1}^L \mathcal{L}_{\text{MoE}}^{(\ell)}.
\label{eq:moe_loss}
\end{equation}

Compared with~\cite{tan2024lightweightvisiontransformer}, our MoE-MAE adopts a lean masked-patch pathway, \textit{i.e.}, masked tokens do not traverse an encoder-like branch, thereby remaining faithful to the original MAE architecture and reducing model parameters.

\subsection{Decoder}
The decoder is lightweight both in depth and dimension when compared to the encoder. In our implementation, it consists of only two MoE transformer encoder layer blocks with embedding dimension $d_{\text{dec}}$, reduced hidden size $d_{\text{dec}}/2$, and three experts per MoE module. Each block reuses the same design as the encoder (GQA and SwiGLU-based MoE with weight sharing), but at smaller scale, keeping the overall cost low.

The decoder input sequence is constructed as follows:  
(i) metadata and class tokens from the encoder are projected into the decoder embedding space,  
(ii) visible patch tokens are projected into the same space, and  
(iii) masked patch positions are filled with a learned mask token.  

This sequence is restored to the original patch order using the stored masking indices and augmented with decoder positional embeddings. After processing by the decoder layers, a linear prediction head maps each token back to pixel values of the corresponding patch.

When compared with~\cite{tan2024lightweightvisiontransformer}, we additionally apply decoder patch-level positional embeddings, which preserves reconstruction fidelity and keeps positional information explicit at the decoder.

\subsection{Pretraining Objective}
The model is trained in a self-supervised manner using a reconstruction-based objective augmented with auxiliary losses. Given an input image $\mathbf{X}$, a random subset of patches $\mathcal{M}$ is masked, while the remaining visible patches $\mathcal{V}$ are passed through the encoder. The decoder reconstructs all patches, but the main loss is applied only to the masked ones:
\begin{equation}
\mathcal{L}_{\text{masked}} = \frac{1}{|\mathcal{M}|} \sum_{i \in \mathcal{M}} \left\| \hat{\mathbf{x}}_i - \mathbf{x}_i \right\|^2 ,
\end{equation}
where $\mathbf{x}_i$ and $\hat{\mathbf{x}}_i$ are the original and decoder prediction for patch $i$, respectively.  

An auxiliary reconstruction term on visible patches is also added:
\begin{equation}
\mathcal{L}_{\text{unmasked}} = \frac{1}{|\mathcal{V}|} \sum_{i \in \mathcal{V}} \left\| \hat{\mathbf{x}}_i - \mathbf{x}_i \right\|^2 .
\end{equation}

Finally, each MoE layer produces a regularization term $\mathcal{L}_{\text{MoE}}^{(\ell)}$ that encourages balanced expert utilization. These are accumulated across all encoder layers, as shown in Eq. \ref{eq:moe_loss}.

The full pretraining objective is thus:
\begin{equation}
\mathcal{L} = \mathcal{L}_{\text{masked}} + \alpha \, \mathcal{L}_{\text{unmasked}} + \beta \, \mathcal{L}_{\text{MoE}},
\end{equation}
where $\alpha$ and $\beta$ are scalar weights controlling the contribution of auxiliary losses. This objective encourages the encoder to learn semantically meaningful patch-level representations while the MoE regularization promotes diverse and balanced expert specialization.

\section{Experiments}
\label{sec:experiments}
\subsection{Datasets}
\label{subsec:datasets}
\paragraph{BigEarthNet-Landsat.} 
\label{par:ben_ls}
The BigEarthNet-Landsat (BEN-LS) dataset~\cite{corley2025landsatbenchdatasetsbenchmarkslandsat} is used for both pretraining and evaluation. BEN-LS is a large-scale multi-label land-cover classification benchmark, originally introduced in~\cite{Sumbul_2019} for Sentinel-2 imagery and later extended to Landsat. It contains 590{,}326 image patches of size $40 \times 40$ pixels, each with seven spectral bands (Coastal/Aerosol, Blue, Green, Red, NIR, SWIR-1, SWIR-2). Each patch is annotated with one or more CORINE Land Cover (CLC) classes, reflecting the heterogeneous composition of land-cover types. The dataset is divided into training, validation, and test splits with 269{,}695, 123{,}723, and 125{,}866 samples, respectively~\cite{corley2025landsatbenchdatasetsbenchmarkslandsat}.

\paragraph{EuroSAT-Landsat.}  
The EuroSAT-Landsat (EuroSAT-LS) dataset~\cite{corley2025landsatbenchdatasetsbenchmarkslandsat} is used to evaluate cross-dataset generalization. It was first introduced in~\cite{helber2019eurosatnoveldatasetdeep} and extracted from Sentinel-2 imagery. The extended EuroSAT-LS contains 27{,}000 image patches of size $22 \times 22$ pixels derived from Landsat imagery, covering 10 land use and land cover classes such as residential, industrial, highway, river, forest, and pasture and containing seven spectral bands (see BEN-LS~\ref{par:ben_ls}). In contrast to BEN-LS, EuroSAT-LS is a single-label classification dataset, where each patch belongs to exactly one class. Furthermore, no explicit metadata (geographic coordinates or acquisition time) is provided, allowing to assess the transferability of representations pretrained with geo-temporal conditioning when such information is unavailable. The training, validation, and test splits are of size 16{,}200, 5{,}400, and 5{,}400, respectively~\cite{corley2025landsatbenchdatasetsbenchmarkslandsat}.

A quick comparison of the two datasets is provided in Table~\ref{tab:datasets}.

\begin{table}[ht]
\centering
\caption{Summary of datasets used for pretraining and evaluation.}
\label{tab:datasets}
\begin{tabular}{lccccc}
\toprule
\textbf{Dataset} & \textbf{Samples} & \textbf{Patch size} & \textbf{Bands} & \textbf{Labels} & \textbf{Metadata} \\
\midrule
BEN-LS~\cite{corley2025landsatbenchdatasetsbenchmarkslandsat,Sumbul_2019} & 590{,}326 & $40 \times 40$ & 7 & Multi-label (CLC) 19 classes & Yes \\
EuroSAT-LS~\cite{helber2019eurosatnoveldatasetdeep} & 27{,}000 & $22 \times 22$ & 7 & 10 classes (single-label) & No \\
\bottomrule
\end{tabular}
\end{table}

\subsection{Experimental Setup}
\label{subsec:experimental_setup}
\paragraph{Hardware.}
All experiments were conducted on a virtualized environment NVIDIA A100 GPU with MIG enabled (MIG 2g.20gb, 20 GB VRAM), corresponding to approximately 25\% of an A100-80GB. The machine was equipped with 8 vCPUs (AMD EPYC 7543) and 62 GB RAM (see Table~\ref{tab:machine_specs}).
\begin{table}[ht]
\centering
\caption{Machine specifications used for all experiments.}
\label{tab:machine_specs}
\begin{tabular}{ll}
\toprule
\textbf{Component} & \textbf{Specification} \\
\midrule
GPU  & NVIDIA A100 (MIG 2g.20gb, 20 GB VRAM, $\sim$25\% of A100-80GB) \\
CPU  & 8 vCPUs (AMD EPYC 7543, 32-Core Processor) \\
RAM  & 62 GB \\
OS   & Linux 5.14.0-570 (x86\_64, KVM virtualized) \\
\bottomrule
\end{tabular}
\end{table}

\paragraph{Pretraining.}  
The MoE-MAE is pretrained on the BEN-LS training split using the self-supervised objective described in Section~\ref{sec:method}. A masking ratio of $75\%$ is used to train the model for $500$ epochs using the AdamW optimizer. A warmup-cosine learning rate scheduler is employed. The learning rate is linearly increased from zero to base value during the first $5\%$ of total training epochs (warmup), after which it decays to a minimum value following a cosine schedule:
\begin{equation}
\label{eq:warmp_cosine_scheduler}
\eta_t = \eta_{\min} + \tfrac{1}{2}(\eta_0 - \eta_{\min})\big(1 + \cos(\pi \cdot p_t)\big),
\end{equation}
where $\eta_0$ is the base learning rate ($0.0003$), $\eta_{\min}$ the final learning rate ($0.0$), and $p_t \in [0,1]$ is the normalized progress after the warmup phase. This schedule has been shown to stabilize training in the early epochs while ensuring smooth convergence.
Loss weights are fixed as $\alpha=0.1$ and $\beta=0.5$ as in~\cite{tan2024lightweightvisiontransformer}. All the pretraining parameters are listed in Table~\ref{tab:pretrain_hyperparams}.

\begin{table}[ht]
\centering
\caption{Pretraining hyperparameters for the proposed MoE-MAE.}
\label{tab:pretrain_hyperparams}
\begin{tabular}{lcc}
\toprule
\textbf{Setting} & \textbf{Value} \\
\midrule
Masking ratio & $75\%$  \\
Epochs & $500$ \\
Batch size & $128$ \\
Optimizer & AdamW  \\
Base learning rate & $0.0003$  \\
Weight decay & $0.05$  \\
Warmup epochs & $0.05 \times$ total epochs \\
Min learning rate & $0.0$ \\
Scheduler & Warmup + cosine decay Eq.~\ref{eq:warmp_cosine_scheduler} \\
Loss weight $\alpha$ & $0.1$ \\
Loss weight $\beta$ & $0.5$ \\
\bottomrule
\end{tabular}
\end{table}

\paragraph{Linear probe evaluation.}
For these experiments, the encoder of the MoE-MAE is frozen and a logistic-regression probe is trained on top of the encoder embeddings, or class tokens using the SAG solver. For multi-label setting, the logistic-regression is wrapped in a One-Vs-Rest scheme. For BEN-LS, the model is trained on the embeddings of the training split, and the results are reported on the test split of the dataset. Since it is a multi-label classification problem, we report micro/macro precision, recall, F1, and micro/macro mAP computed from predicted probabilities (see Table~\ref{tab:probe_settings}).

\paragraph{Cross-dataset transfer.}  
To assess generalization across datasets, we evaluate the pretrained encoder on EuroSAT-LS. Unlike BEN-LS, EuroSAT-LS does not provide geo-temporal metadata; hence, only imagery tokens are available at train and test time. We train a logistic-regression probe on frozen embeddings using the same setup as described above, considering both the class token alone and all embeddings. Since EuroSAT-LS is a single-label classification dataset, the probe is trained with cross-entropy loss, and we report overall accuracy as well as micro/macro precision, recall, and F1. This evaluation highlights that the learned representations remain effective even when metadata is absent and the label space differs from pretraining (see Table~\ref{tab:probe_settings}).

\begin{table}[ht]
\centering
\caption{Training configurations for linear probe and cross-dataset transfer experiments.}
\label{tab:probe_settings}
\begin{tabular}{lcc}
\toprule
 & \textbf{BigEarthNet-LS (linear probe)} & \textbf{EuroSAT-LS (cross-dataset)} \\
\midrule
Input embeddings & CLS only / all tokens & CLS only / all tokens \\
Metadata & Used in pretraining, available & Used in pretraining, \emph{absent} at test time \\
Task type & Multi-label classification & Single-label classification \\
Probe model & Logistic Regression (One-vs-Rest) & Logistic Regression \\
Solver / max iter & SAG / 1000 & SAG / 1000 \\
Loss & BCE with logits (per class) & Cross-entropy \\
Metrics & mAP (micro/macro), P/R/F1 (micro/macro) & Accuracy, P/R/F1 (micro/macro) \\
\bottomrule
\end{tabular}
\end{table}

\subsection{Pretraining on BEN-LS}
\begin{figure}[ht]
\centering
\begin{subfigure}{0.48\linewidth}
    \centering
    \includegraphics[width=\linewidth]{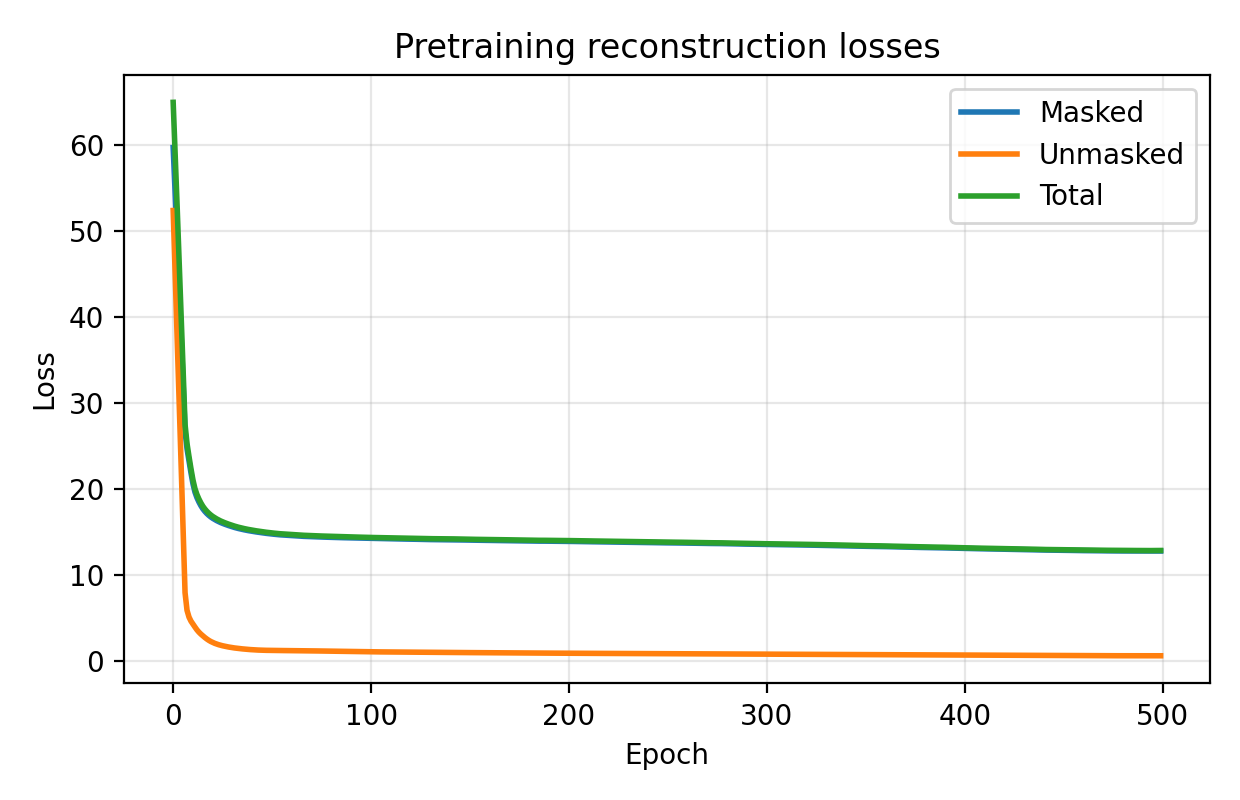}
    \caption{Reconstruction losses}
\end{subfigure}\hfill
\begin{subfigure}{0.48\linewidth}
    \centering
    \includegraphics[width=\linewidth]{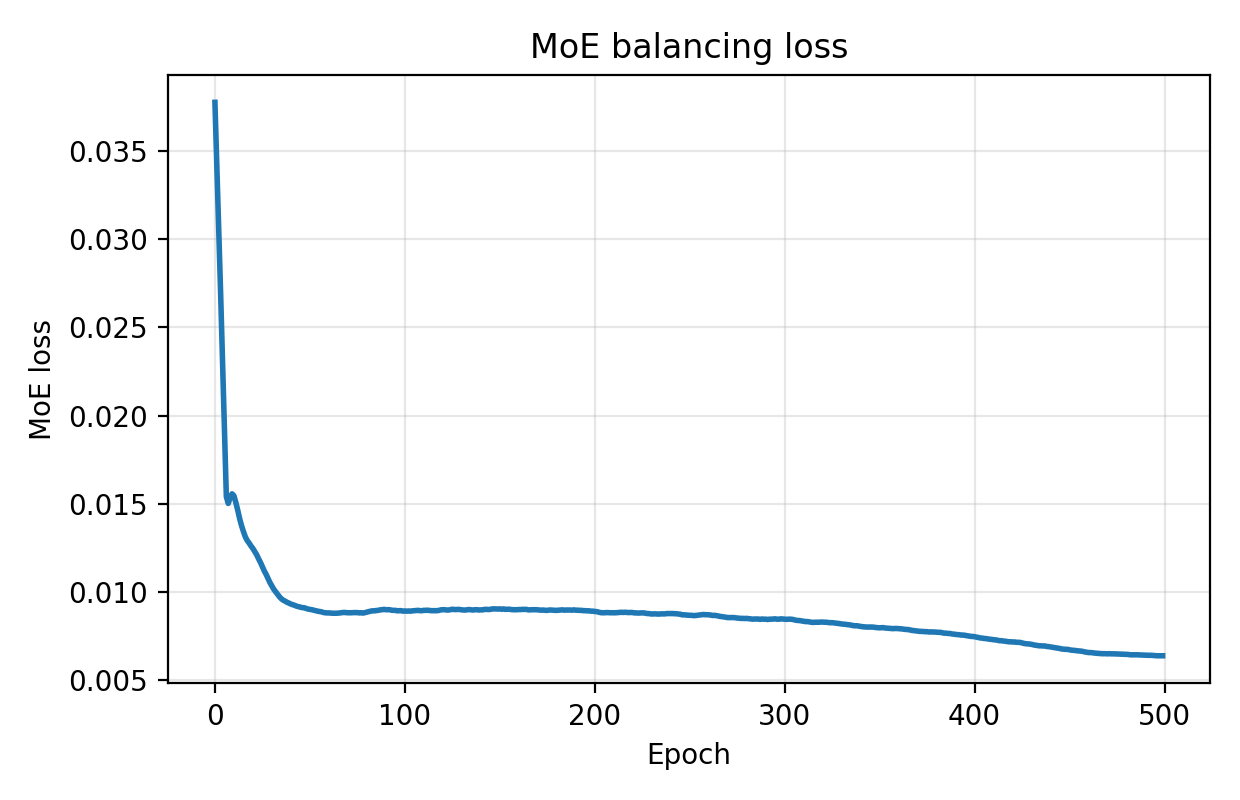}
    \caption{MoE balancing loss}
\end{subfigure}

\vspace{0.35cm}
\begin{subfigure}{0.7\linewidth}
    \centering
    \includegraphics[width=\linewidth]{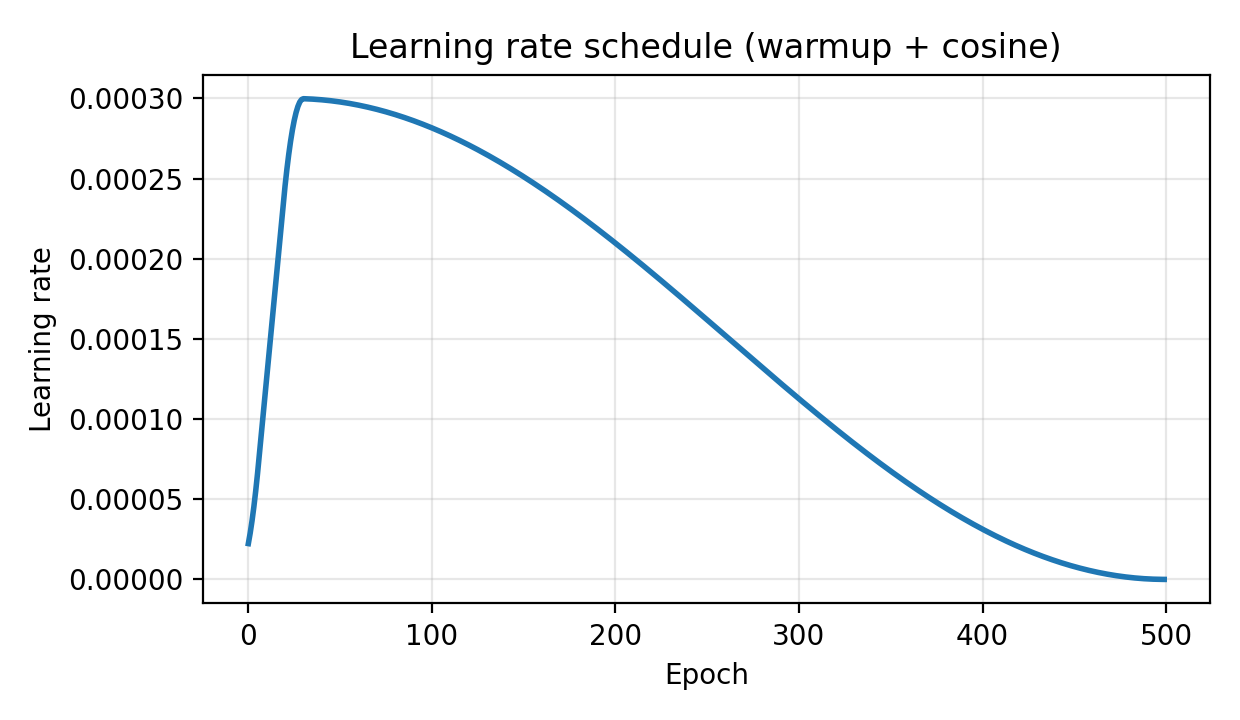}
    \caption{Learning rate schedule}
\end{subfigure}

\caption{Pretraining dynamics of MoE-MAE on BEN-LS.}
\label{fig:pretrain_dynamics}
\end{figure}

We first analyze the behavior of the proposed MoE-MAE during pretraining on the BEN-LS dataset. 
Fig.~\ref{fig:pretrain_dynamics} summarizes the training dynamics. Reconstruction losses (the masked, unweighted auxiliary unmasked, and total losses) are depicted in Fig.~\ref{fig:pretrain_dynamics}(a). They steadily decrease across epochs, indicating that the encoder learns semantically meaningful representations suitable for reconstruction. Fig.~\ref{fig:pretrain_dynamics}(b) reports the MoE balancing loss, which remains bounded and decreases smoothly. This shows that experts are consistently utilized without collapse, and confirms that NoisyTop-$k$ routing combined with the load-balancing regularizer achieves stable expert routing throughout training. Finally, Fig.~\ref{fig:pretrain_dynamics}(c) illustrates the learning rate schedule, which consists of a linear warmup phase during the first $5\%$ of training epochs followed by cosine decay. 

Overall, these results demonstrate that the compact MoE-MAE converges stably on BEN-LS, effectively balancing masked reconstruction and MoE regularization objectives.

\begin{figure}[ht]
\centering
\begin{minipage}{0.3\linewidth}
    \centering
    \textbf{Original}
\end{minipage}\hfill
\begin{minipage}{0.3\linewidth}
    \centering
    \textbf{Full reconstruction}
\end{minipage}\hfill
\begin{minipage}{0.3\linewidth}
    \centering
    \textbf{Masked (75\%)}
\end{minipage}

\vspace{0.2cm}
\begin{minipage}{0.3\linewidth}
    \centering
    \includegraphics[width=\linewidth]{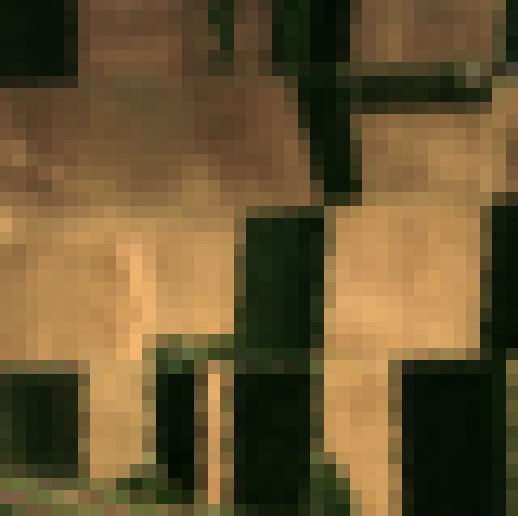}
\end{minipage}\hfill
\begin{minipage}{0.3\linewidth}
    \centering
    \includegraphics[width=\linewidth]{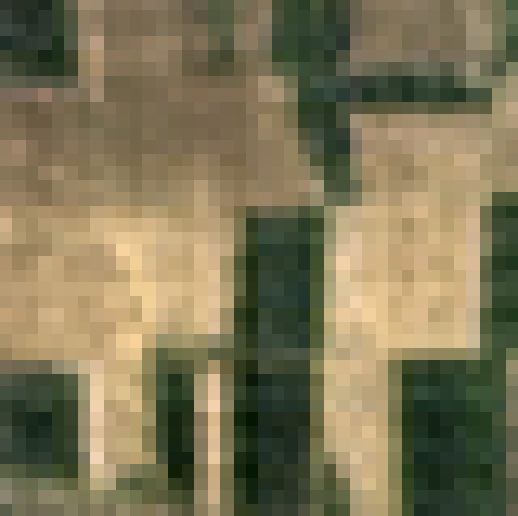}
\end{minipage}\hfill
\begin{minipage}{0.3\linewidth}
    \centering
    \includegraphics[width=\linewidth]{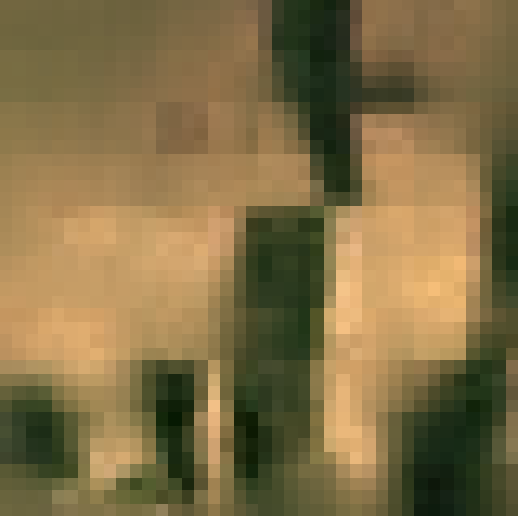}
\end{minipage}

\vspace{0.2cm}
\begin{minipage}{0.3\linewidth}
    \centering
    \includegraphics[width=\linewidth]{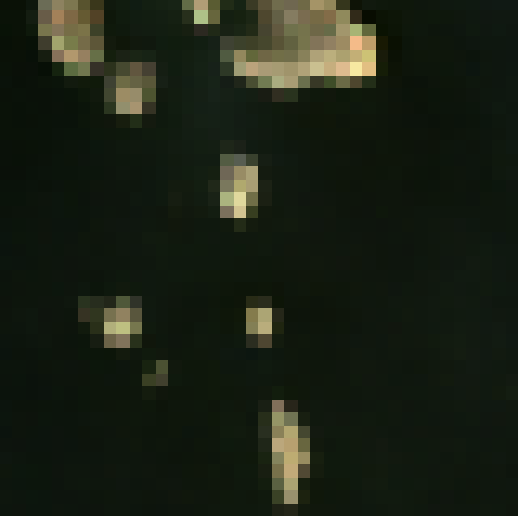}
\end{minipage}\hfill
\begin{minipage}{0.3\linewidth}
    \centering
    \includegraphics[width=\linewidth]{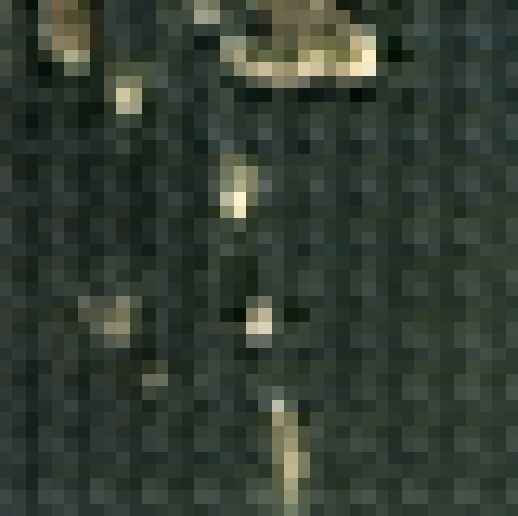}
\end{minipage}\hfill
\begin{minipage}{0.3\linewidth}
    \centering
    \includegraphics[width=\linewidth]{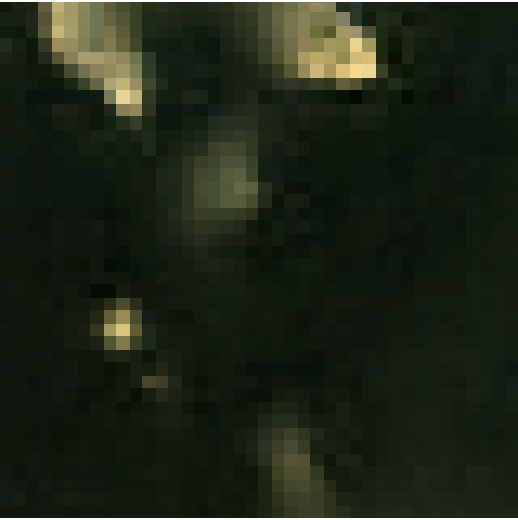}
\end{minipage}

\vspace{0.2cm}
\begin{minipage}{0.3\linewidth}
    \centering
    \includegraphics[width=\linewidth]{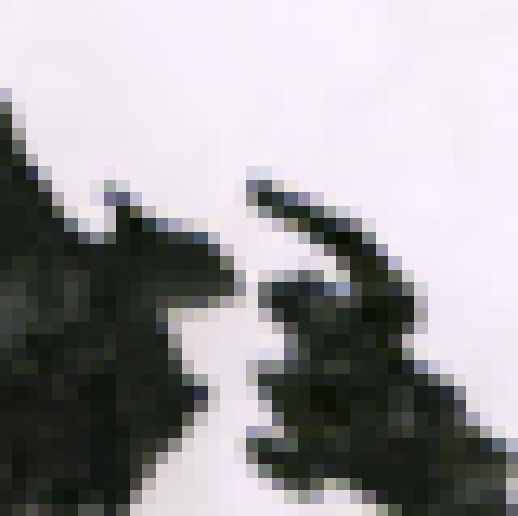}
\end{minipage}\hfill
\begin{minipage}{0.3\linewidth}
    \centering
    \includegraphics[width=\linewidth]{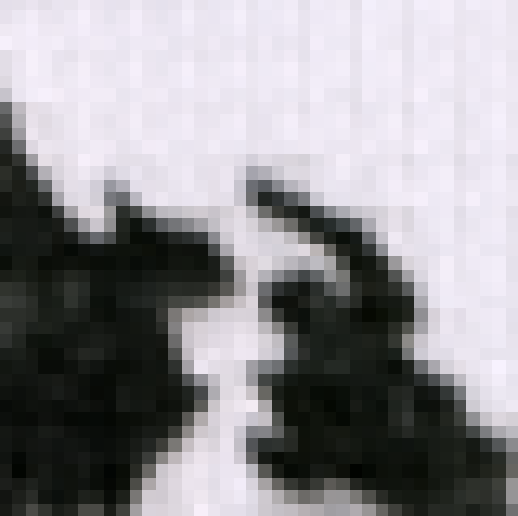}
\end{minipage}\hfill
\begin{minipage}{0.3\linewidth}
    \centering
    \includegraphics[width=\linewidth]{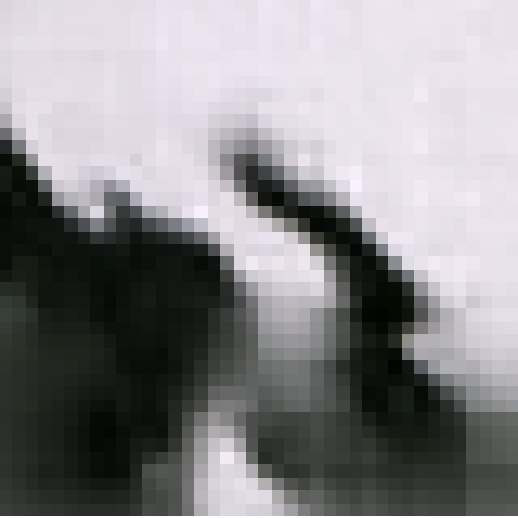}
\end{minipage}

\caption{Qualitative reconstruction examples on BEN-LS. Each row shows a different sample: original input (left), reconstruction without masking (middle), and reconstruction from 75\% masked patches (right).}
\label{fig:recon_examples}
\end{figure}

Qualitative reconstruction results are depicted in Fig.~\ref{fig:recon_examples}. Each row corresponds to one BEN-LS sample from the test split. Columns show the original image, the reconstruction without masking, and the reconstruction when $75\%$ of patches are masked, respectively. These results show that even under heavy masking, the model manages to recover spatial and spectral structures. This confirms that the compact MoE encoder learns semantically meaningful representations.
 
\subsection{Linear Probing on BEN-LS}

We first evaluate our model on the BigEarthNet-LS multi-label test split using a logistic regression probe. The classification metrics for the CLS tokens and all encoder tokens and average pooling of all tokens are reported in Table~\ref{tab:linear_probe_benls}. Classification with all tokens achieves better performance than classification using CLS tokens, and average pooling. This indicates that these features capture compact global semantics, and can be better adapted for image classification problems.

\begin{table}[ht]
\centering
\caption{Linear probe performance on BEN-LS (multi-label test split). Results are reported for using only the CLS token versus all encoder tokens as features.}
\label{tab:linear_probe_benls}
\begin{tabular}{lcccccc}
\toprule
\textbf{Embedding} & \textbf{Precision (micro)} & \textbf{Recall (micro)} & \textbf{F1 (micro)} & \textbf{F1 (macro)} & \textbf{mAP (micro)} & \textbf{mAP (macro)} \\
\midrule
CLS token   & 0.755 & 0.551 & 0.638 & 0.450 & 0.748 & 0.582 \\
All tokens  & 0.737 & \textbf{0.615} & \textbf{0.670} & \textbf{0.540} & \textbf{0.767} & \textbf{0.600} \\
All tokens (Avg.) & \textbf{0.763} & 0.514 & 0.614 & 0.419 & 0.731 & 0.557\\
\bottomrule
\end{tabular}
\end{table}

The obtained results are further compared against established baselines from LandsatBench~\cite{corley2025landsatbenchdatasetsbenchmarkslandsat}, shown in Table~\ref{tab:landsatbench}. 
Our all-token mAP of 0.767 outperforms SSL4EO-L ResNet-50 MoCo v2 (0.761) and approaches SSL4EO-L ViT-S/16 MoCo v2 (0.775), despite being trained only on BEN-LS and containing just $\sim$2.3M parameters. 
This highlights the efficiency of metadata-aware MoE-MAEs for learning transferable EO representations.

\subsection{Cross-Dataset Transfer to EuroSAT-LS}
We next evaluate transferability to EuroSAT-LS, a single-label land-cover dataset with different label semantics than BEN-LS that has no metadata (geographical coordinates and acqusition dates). 
Table~\ref{tab:linear_probe_eurosat} shows that all-token embeddings achieve 84.2\% accuracy, outperforming CLS-token and average pooling token embeddings (78.4\% and 74.3\%, respectively). 
This suggests that the learned representations are transferable across datasets even when they are missing metadata.

\begin{table}[ht]
\centering
\caption{Linear probe performance on EuroSAT-LS.}
\label{tab:linear_probe_eurosat}
\begin{tabular}{lcccc}
\toprule
\textbf{Embedding} & \textbf{OA\%}& \textbf{Precision (macro)} &\textbf{Recall (macro)} & \textbf{F1 (macro)}\\
\midrule
CLS token   & 78.4 &0.781 &0.779 &0.777  \\
All tokens  & \textbf{84.2} & \textbf{0.845} & \textbf{0.838} & \textbf{0.839}\\
All tokens (Avg.) & 74.3 & 0.739 & 0.737 & 0.731\\
\bottomrule
\end{tabular}
\end{table}

We again compare with external baselines from LandsatBench (Table~\ref{tab:landsatbench}). 
Our CLS-token accuracy of 84.2\% outperforms many larger models that are trained on more data. However, it falls behind some bigger models.
This demonstrates that even when trained solely on BEN-LS, our compact MoE-MAE transfers competitively to new datasets without requiring large-scale multi-sensor pretraining. It also suggests that our MoE-MAE can still benefit from further pretraining on larger dataset to capture more image semantics. 

\begin{table}[ht]
\centering
\caption{Comparison with pretrained baselines on EuroSAT-LS (OA) and BigEarthNet-LS (mAP micro). 
All results use logistic regression probes following the LandsatBench protocol~\cite{corley2025landsatbenchdatasetsbenchmarkslandsat}. 
Dashes indicate results not reported in the original benchmark. 
Values in {\color{red}red} are lower than our MoE-MAE (CLS) results.}
\label{tab:landsatbench}
\begin{tabular}{lcc}
\toprule
\textbf{Model} & \textbf{EuroSAT-L (LP)} & \textbf{BigEarthNet-LS (LP)} \\
\midrule
ImageNet ResNet-18         & {\color{red}81.0} & {\color{red}70.2} \\
ImageNet ResNet-50         & {\color{red}79.0} & {\color{red}71.9} \\
ImageNet ViT-S/16          & \textbf{86.4} & {\color{red}74.9} \\
SSL4EO-L ResNet-18 SimCLR & {\color{red}68.5} & {\color{red}60.3} \\
SSL4EO-L ResNet-18 MoCo v2 & {\color{red}82.8} & {\color{red}74.0} \\
SSL4EO-L ResNet-50 SimCLR & {\color{red}73.5} & {\color{red}63.1} \\
SSL4EO-L ResNet-50 MoCo v2 & {\color{red}82.5} & {\color{red}76.1} \\
SSL4EO-L ViT-S/16 SimCLR  & {\color{red}77.6} & {\color{red}68.3} \\
SSL4EO-L ViT-S/16 MoCo v2 & \textbf{88.8} & \textbf{77.5} \\
DOFA-B/16                  & \textbf{87.3} & --   \\
Satlas Swin V2-B           & \textbf{89.6} & --   \\
DOFA-L/16                  & \textbf{89.8} & --   \\
Prithvi-EO-100M            & \textbf{89.4} & --   \\
Prithvi-EO-2.0 300M        & \textbf{91.1} & --   \\
Prithvi-EO-2.0 600M        & \textbf{91.9} & --   \\
\midrule
\textbf{MoE-MAE (CLS token)} & 78.4 & 74.8 \\
\textbf{MoE-MAE (All tokens)} & 84.2 & \textbf{76.7} \\
\textbf{MoE-MAE (All tokens Avg.)} &74.3 & 73.1\\
\bottomrule
\end{tabular}
\end{table}

\begin{figure}[ht]
\centering
\begin{subfigure}{0.48\linewidth}
    \centering
    \includegraphics[width=\linewidth]{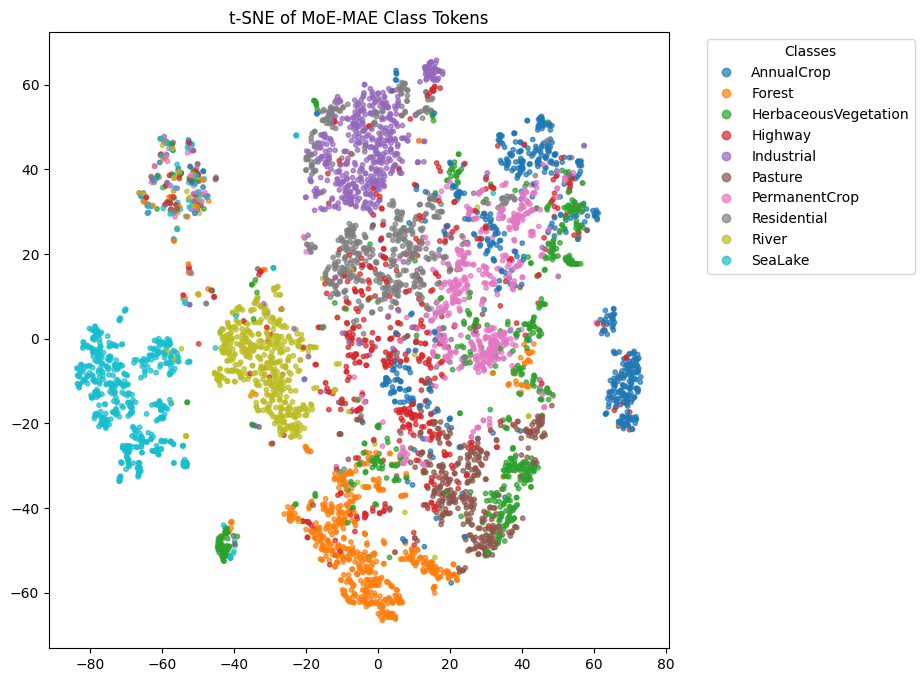}
    \caption{CLS token}
\end{subfigure}\hfill
\begin{subfigure}{0.48\linewidth}
    \centering
    \includegraphics[width=\linewidth]{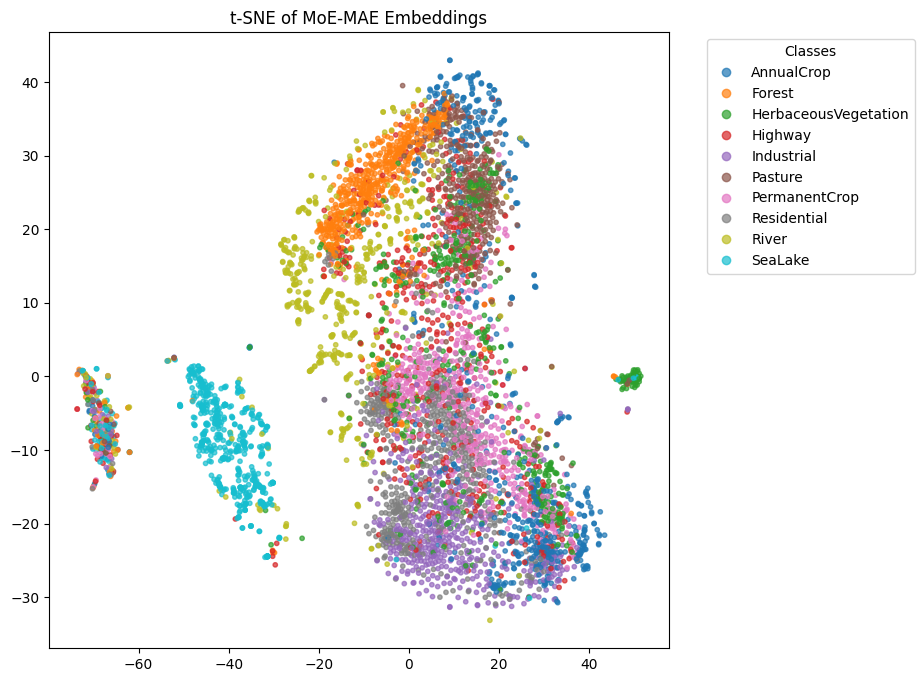}
    \caption{All tokens}
\end{subfigure}
\caption{t-SNE (2D) of EuroSAT-LS embeddings extracted from the frozen encoder. Colors denote class labels.}
\label{fig:eurosat_tsne}
\end{figure}

Fig.~\ref{fig:eurosat_tsne} illustrates the geometry of EuroSAT-LS features. 
The CLS token yields tighter, well-separated clusters (e.g., \emph{SeaLake}, \emph{River}), consistent with its stronger cross-dataset accuracy. 
All-token embeddings expose additional intra-class structure (useful for fine-grained cues) but show slightly more overlap between semantically related classes.

\subsection{Compactness vs. Scale}
A key motivation of this work is to investigate whether small-scale architectures can serve as practical foundations for EO representation learning. Recent EO transformer-based models, \textit{e.g.,} DOFA, Satlas, Clay, and Prithvi contains tens to hundreds of millions parameters, with the largest Prithvi variant containing 600M (see Table~\ref{tab:model_params}) compared to $\sim$2.3M in the MoE-MAE encoder.
Looking at the results in Tables~\ref{tab:linear_probe_benls} and~\ref{tab:model_params}, one can notice the following: MoE-MAE achieves competitive performance on both BEN-LS and EuroSAT-LS, despite its compact size. Additionally, in some cases, it outperforms baselines trained on ImageNet or SSL4EO-L. 
This demonstrates that geo-temporal metadata conditioning combined with sparse MoE routing can significantly boost the representational capacity of lightweight models.  

Using lightweight models has several practical implications. First, compact architectures lower the barrier to pretraining in EO, making work more reproducible and accessible in resource-constrained environments. Second, smaller models are easier to deploy at scale, enabling on-device use and near-real-time inference. Finally, the results suggest that scaling compact, metadata-aware MoE-MAEs on larger and more diverse EO datasets could be a cost-effective path to general-purpose EO foundation models.

\begin{table}[ht]
\centering
\caption{Model size comparison. Our MoE-MAE is two orders of magnitude smaller than recent EO foundation models.}
\label{tab:model_params}
\begin{tabular}{lcc}
\toprule
\textbf{Model} & \textbf{Type} & \textbf{Parameters} \\
\midrule
ResNet-18 (ImageNet / SSL4EO-L) & CNN & $\sim$11M \\
ResNet-50 (ImageNet / SSL4EO-L) & CNN & $\sim$25M \\
ViT-S/16 (ImageNet / SSL4EO-L) & Transformer & $\sim$22M \\
DOFA-B/16~\cite{xiong2024neuralplasticityinspiredmultimodalfoundation} & Transformer & $\sim$111M \\
DOFA-L/16~\cite{xiong2024neuralplasticityinspiredmultimodalfoundation} & Transformer & $\sim$330M \\
Satlas Swin V2-B~\cite{bastani2023satlaspretrainlargescaledatasetremote} & Transformer & $\sim$88M \\
Prithvi-EO-100M~\cite{jakubik2023foundationmodelsgeneralistgeospatial} & Transformer & 100M \\
Prithvi-EO-2.0 300M~\cite{szwarcman2025prithvieo20versatilemultitemporalfoundation} & Transformer & 300M \\
Prithvi-EO-2.0 600M~\cite{szwarcman2025prithvieo20versatilemultitemporalfoundation} & Transformer & 600M \\
\midrule
\textbf{Ours: MoE-MAE (CLS/All tokens)} & Transformer (MoE) & \textbf{~2.5M} \\
\bottomrule
\end{tabular}
\end{table}

\subsection{MoE Ablation Analysis}
To better understand the behavior of the MoE layers, a qualitative and quantitative ablation study is conducted. This analysis focuses on expert routing, contribution, and importance.

\paragraph{Deterministic routing.}
For analysis, the stochastic noise in NoisyTop-$k$ gating is disabled, to ensure reproducible and consistent expert assignments across runs.

\paragraph{Contribution maps.}
We compute per-expert contribution maps by measuring the activation norm of gated expert outputs for every patch token:
\begin{equation}
C_{i,e} = \| g_{i,e} \cdot f_e(\mathbf{z}_i) \|_2 ,
\end{equation}
where $f_e$ is the expert $e$, $g_{i,e}$ is the weight of the gating, and $\mathbf{z}_i$ the embedding of the token. These maps indicate which spatial regions each expert is most responsible for, and can be overlaid on the RGB image for interpretation.

\paragraph{Ablation maps.}
To quantify expert importance, we disable one expert at a time during inference and propagate the tokens through the remainder of the encoder. The effect is measured as the difference in final patch embeddings:
\begin{equation}
\Delta_{i,e} = \| \mathbf{y}_i - \mathbf{y}^{(-e)}_i \|_2 ,
\end{equation}
where $\mathbf{y}_i$ is the normal output and $\mathbf{y}^{(-e)}_i$ the output when expert $e$ is ablated. Large values of $\Delta_{i,e}$ indicate critical contributions of expert $e$ for token $i$.

\begin{figure}[ht]
    \centering
    \begin{tabular}{ccc}
        \includegraphics[width=0.235\linewidth]{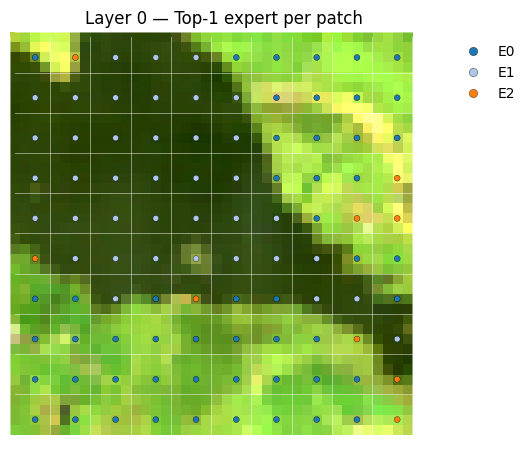} &
        \includegraphics[width=0.30\linewidth]{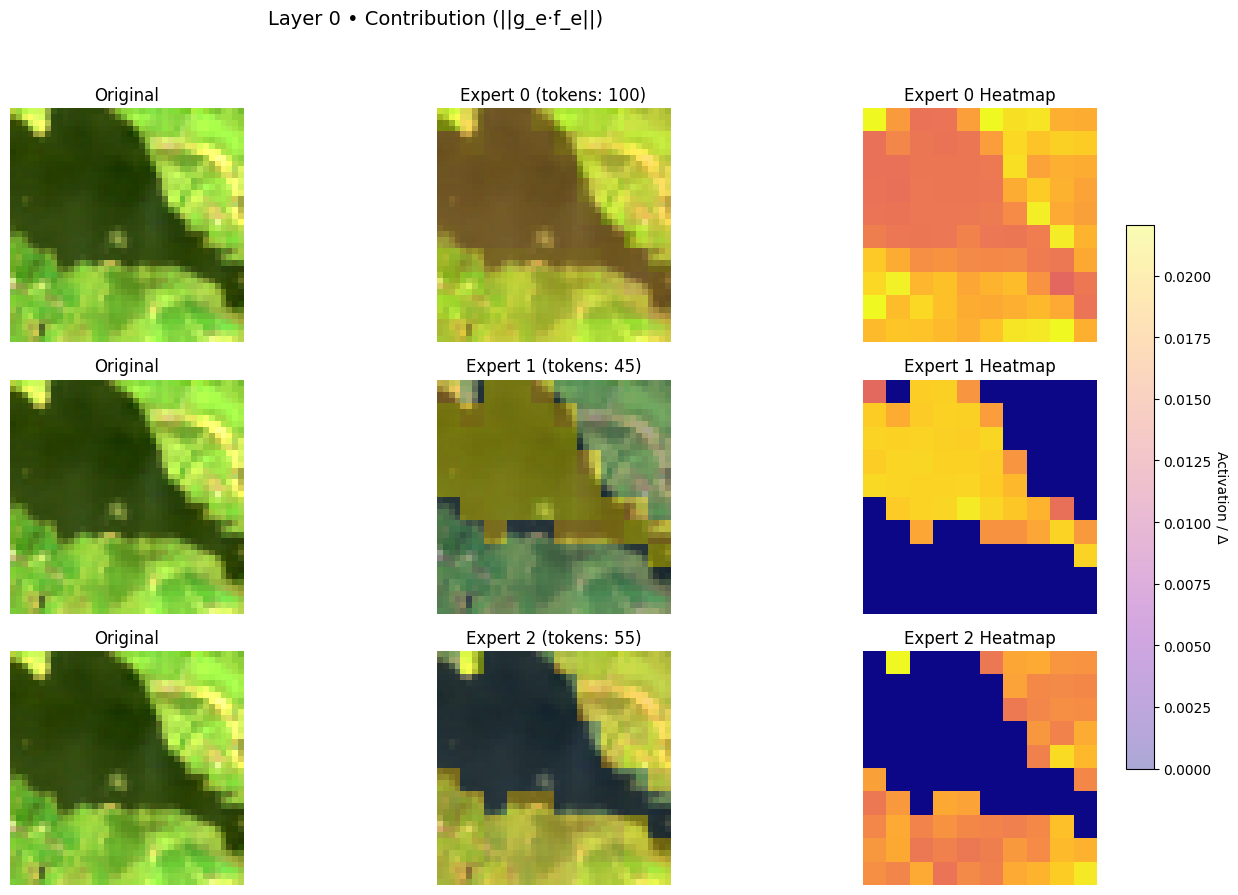} &
        \includegraphics[width=0.30\linewidth]{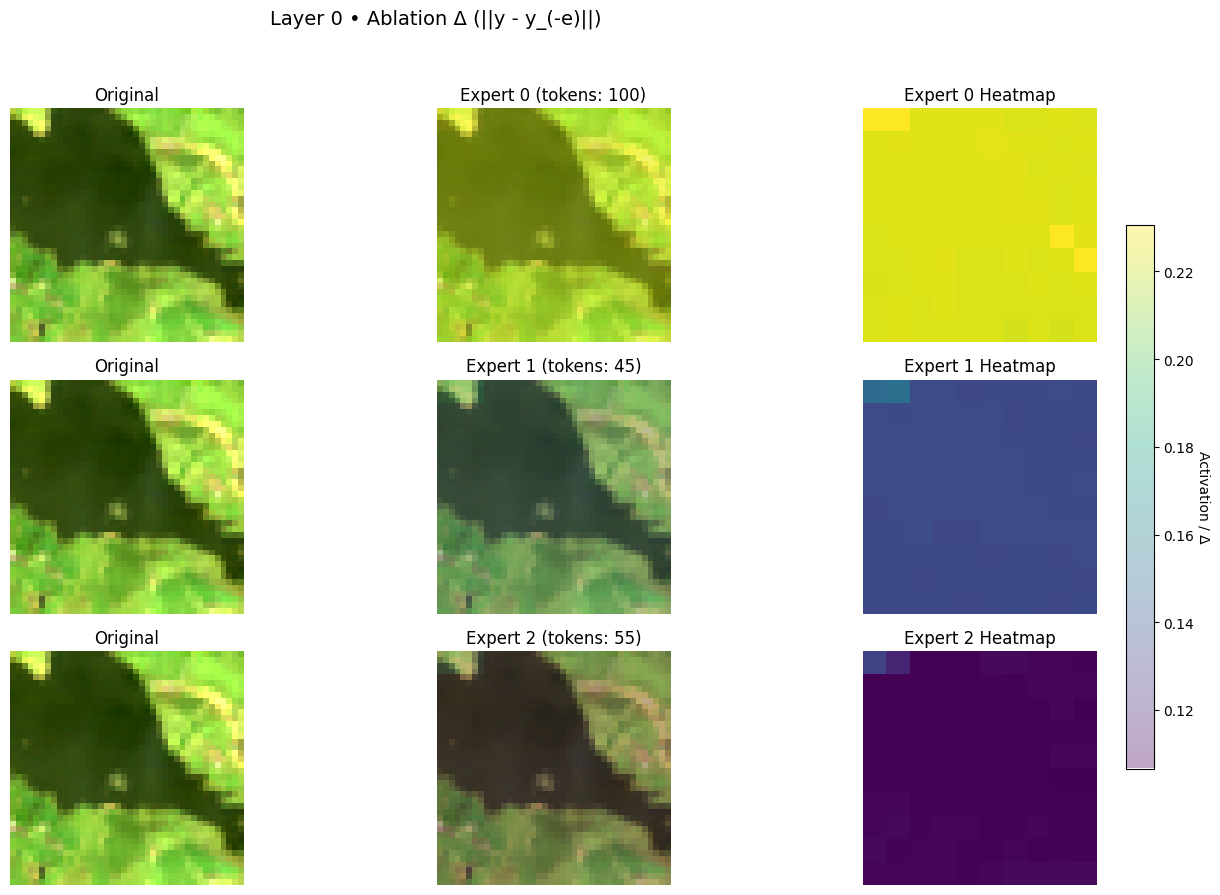} \\
        \includegraphics[width=0.235\linewidth]{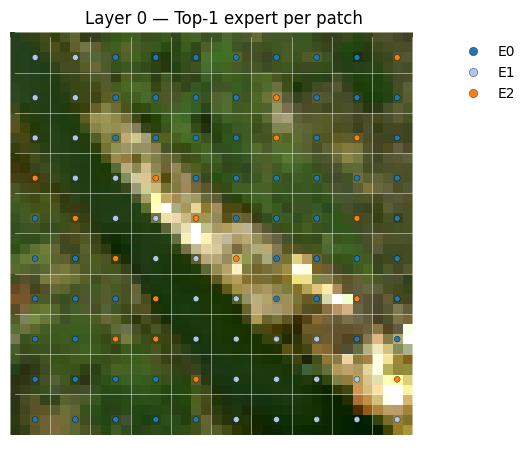} &
        \includegraphics[width=0.30\linewidth]{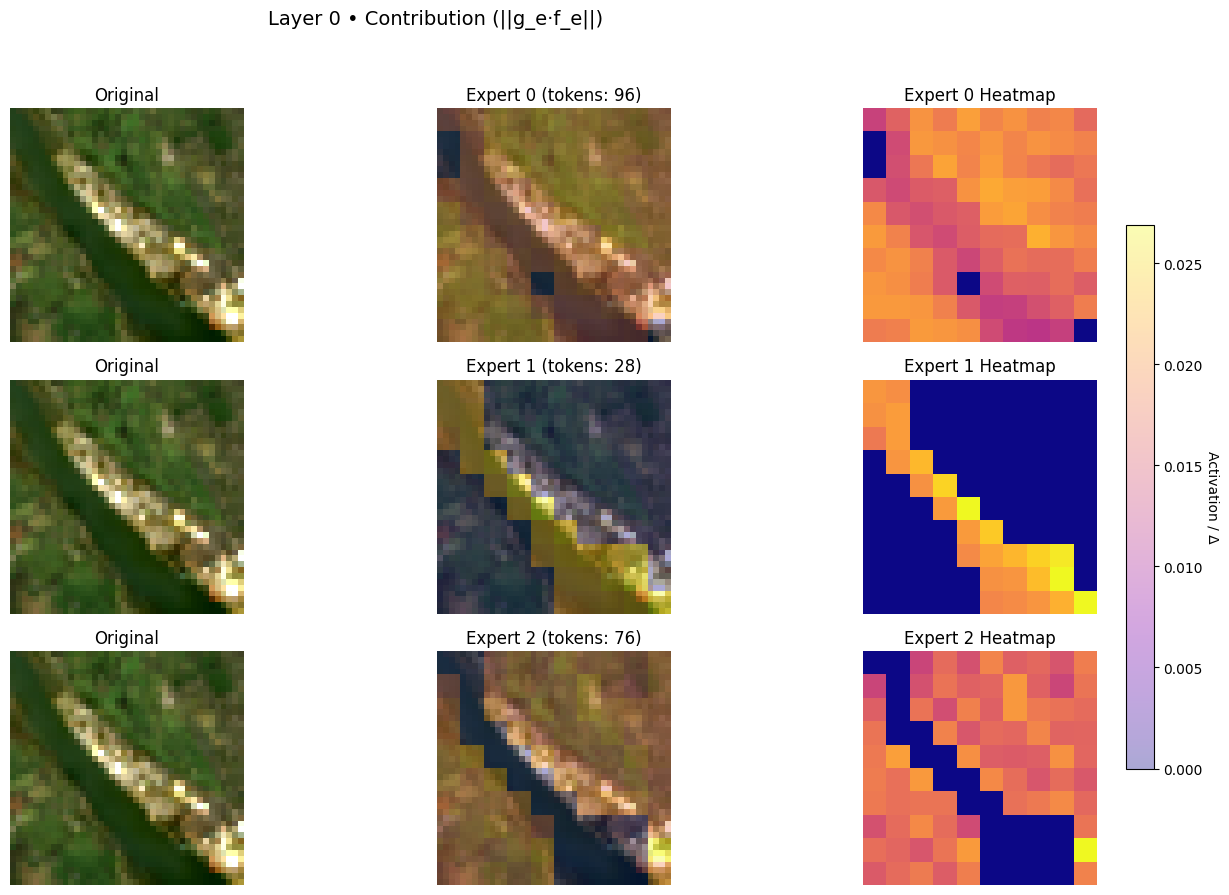} &
        \includegraphics[width=0.30\linewidth]{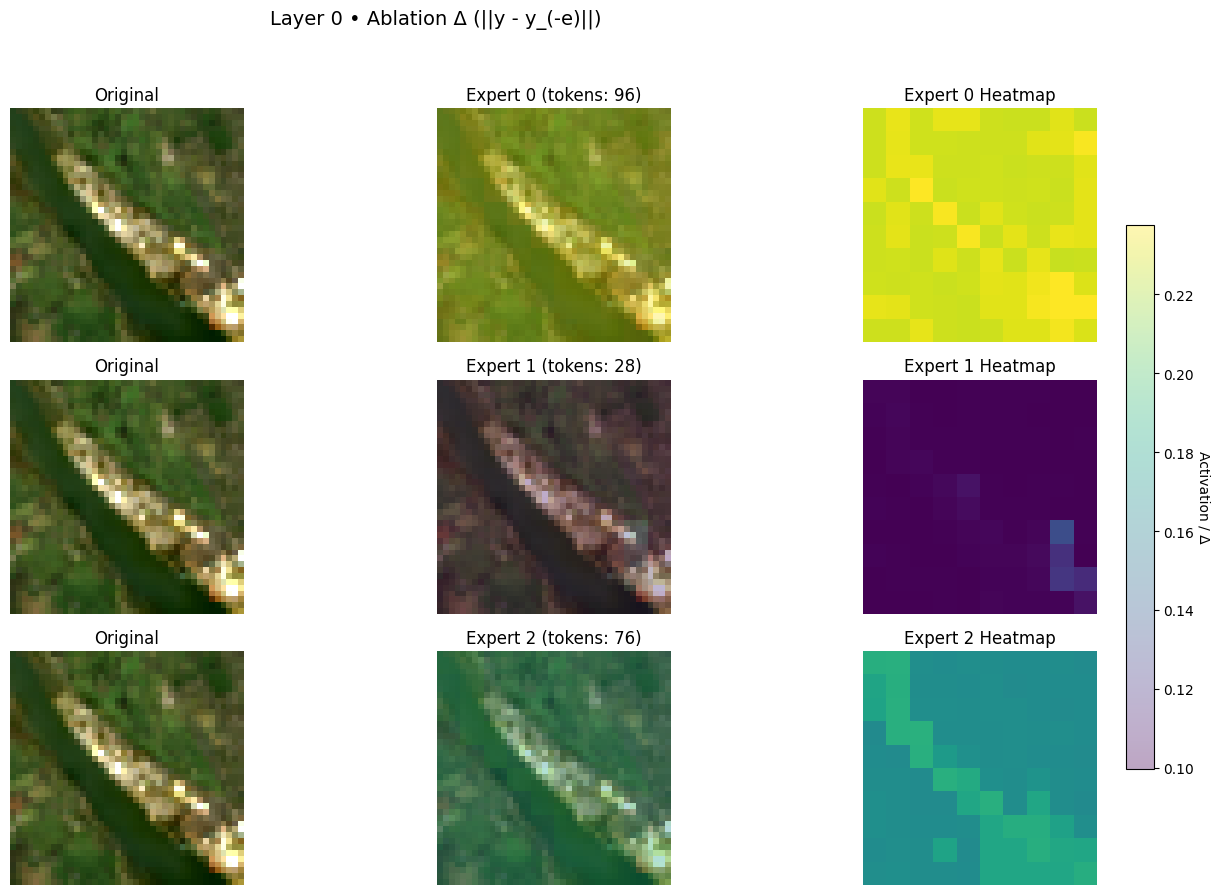} \\
        \includegraphics[width=0.235\linewidth]{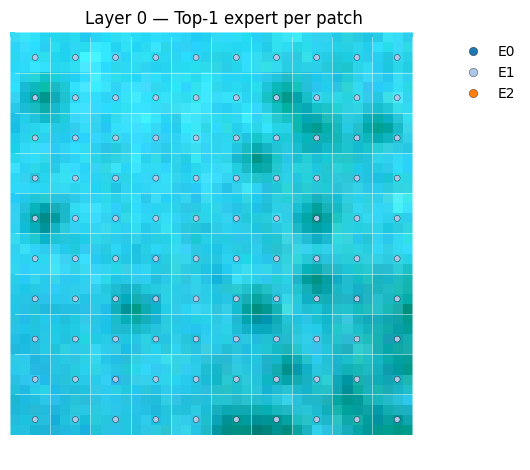} &
        \includegraphics[width=0.30\linewidth]{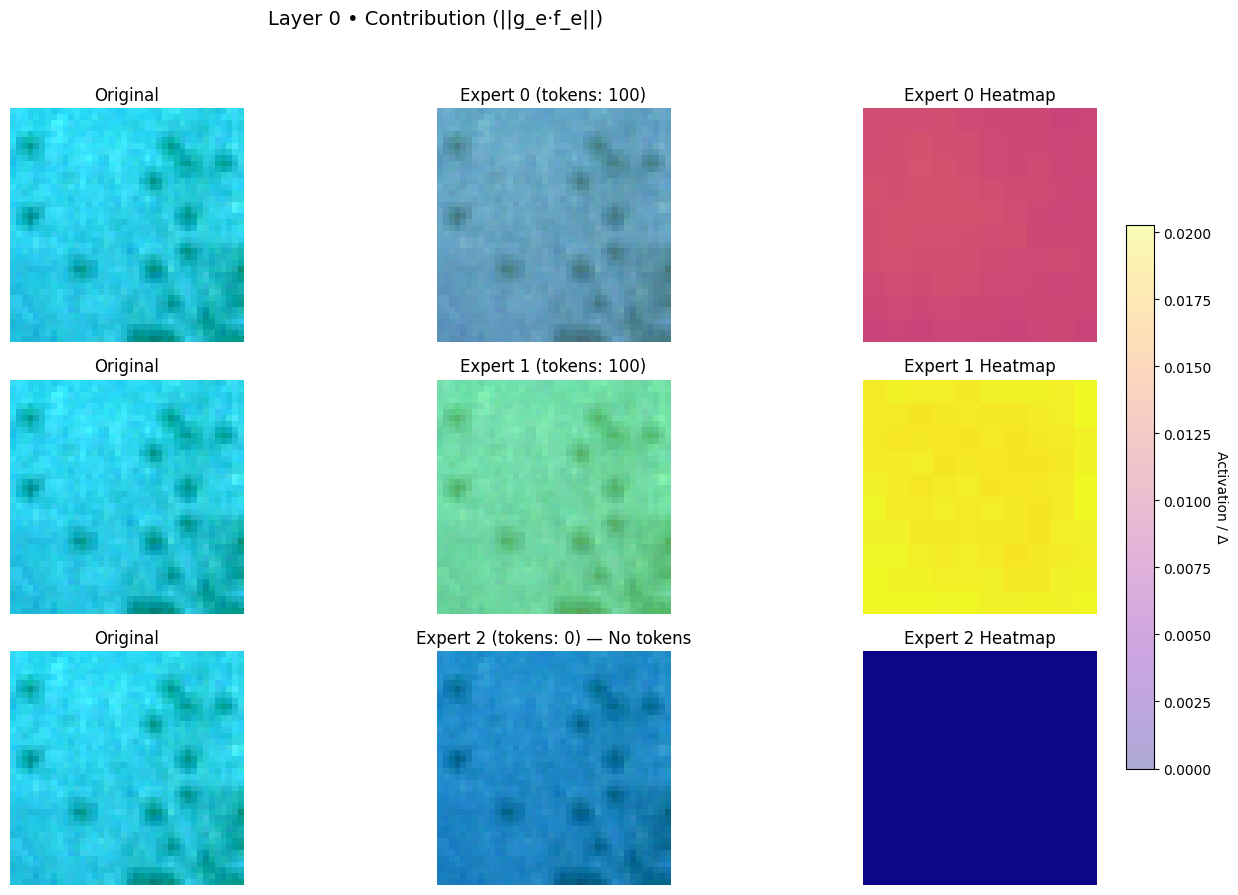} &
        \includegraphics[width=0.30\linewidth]{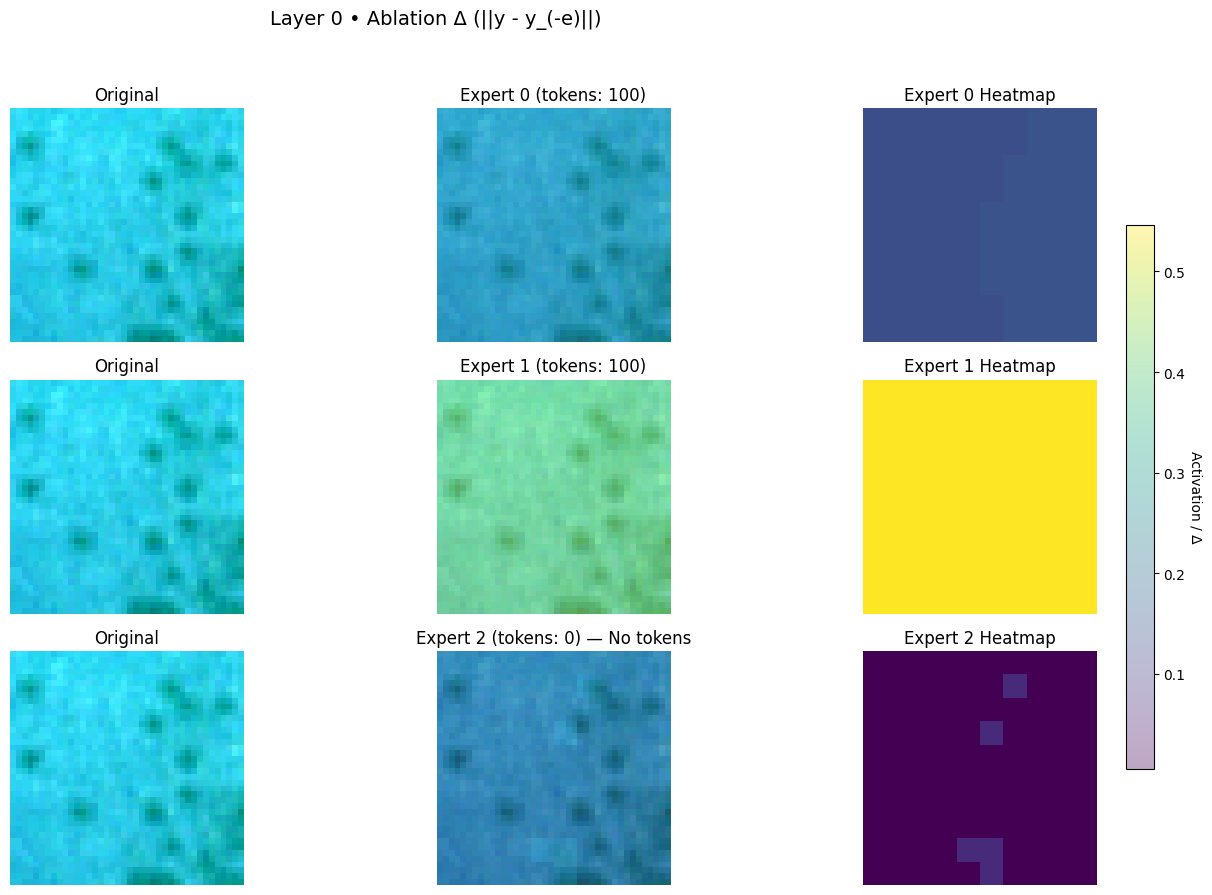} \\
        \includegraphics[width=0.235\linewidth]{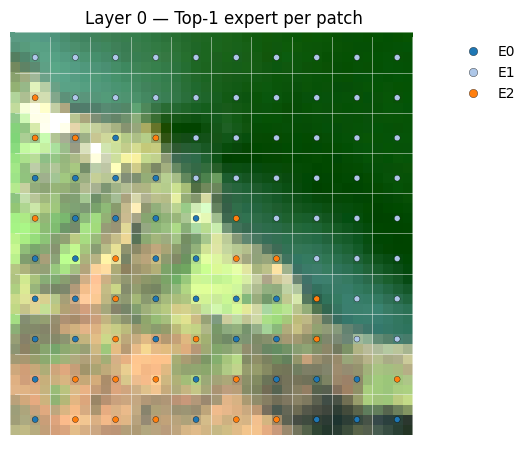} &
        \includegraphics[width=0.30\linewidth]{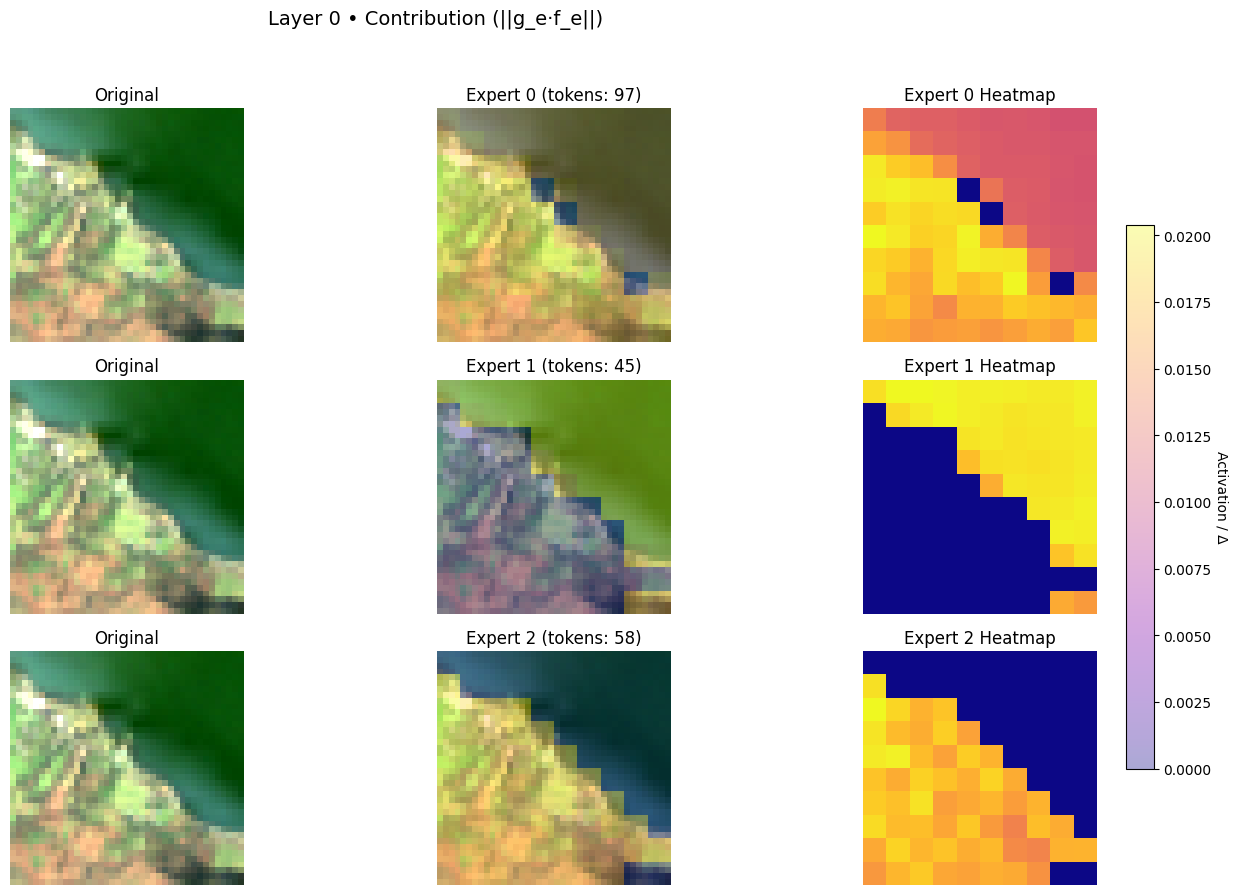} &
        \includegraphics[width=0.30\linewidth]{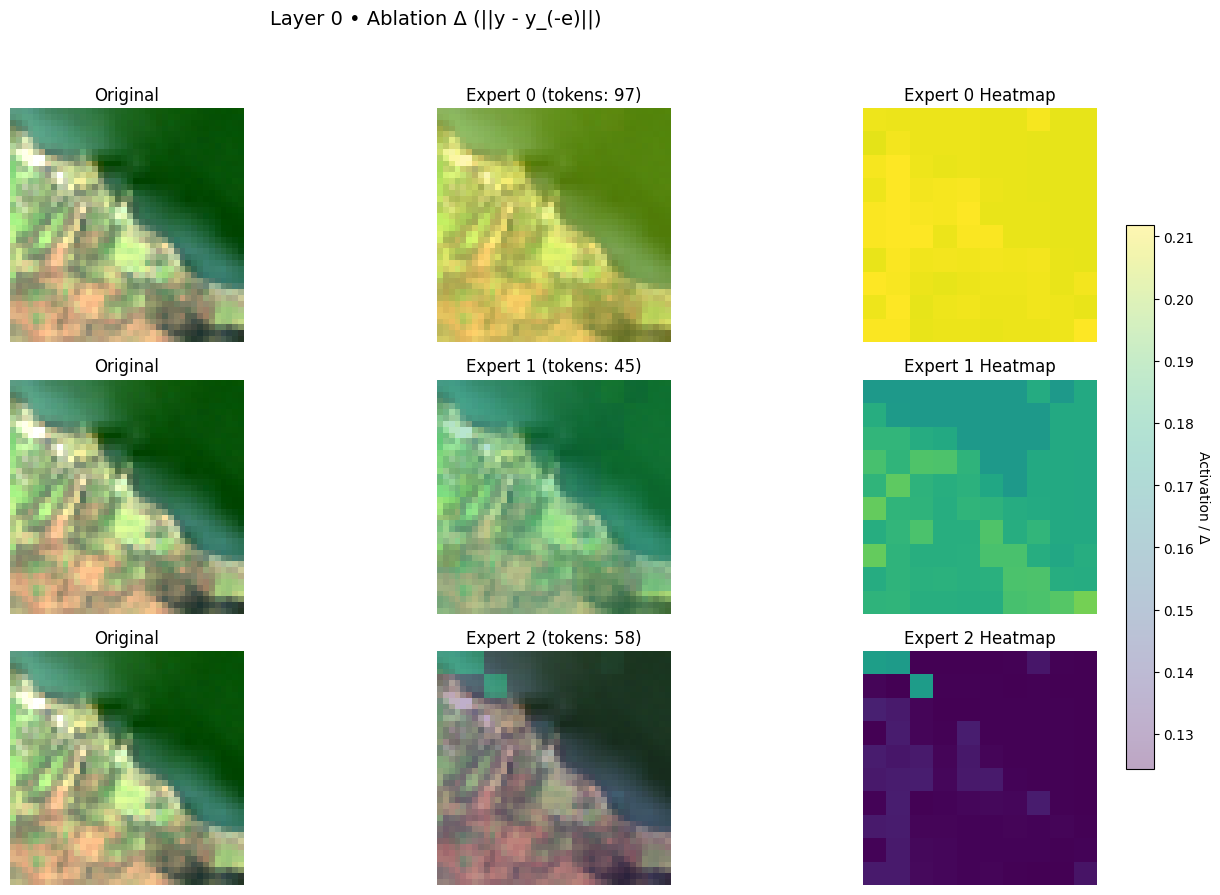} \\
        \includegraphics[width=0.235\linewidth]{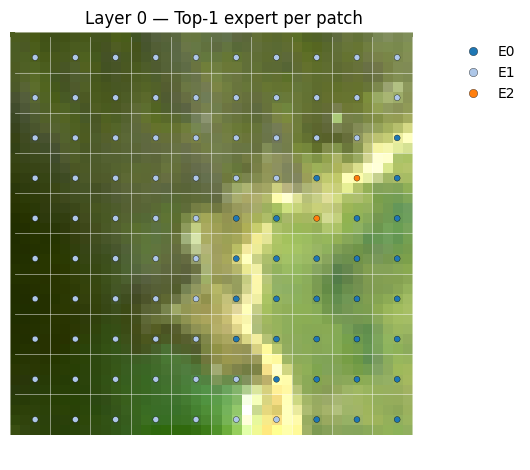} &
        \includegraphics[width=0.30\linewidth]{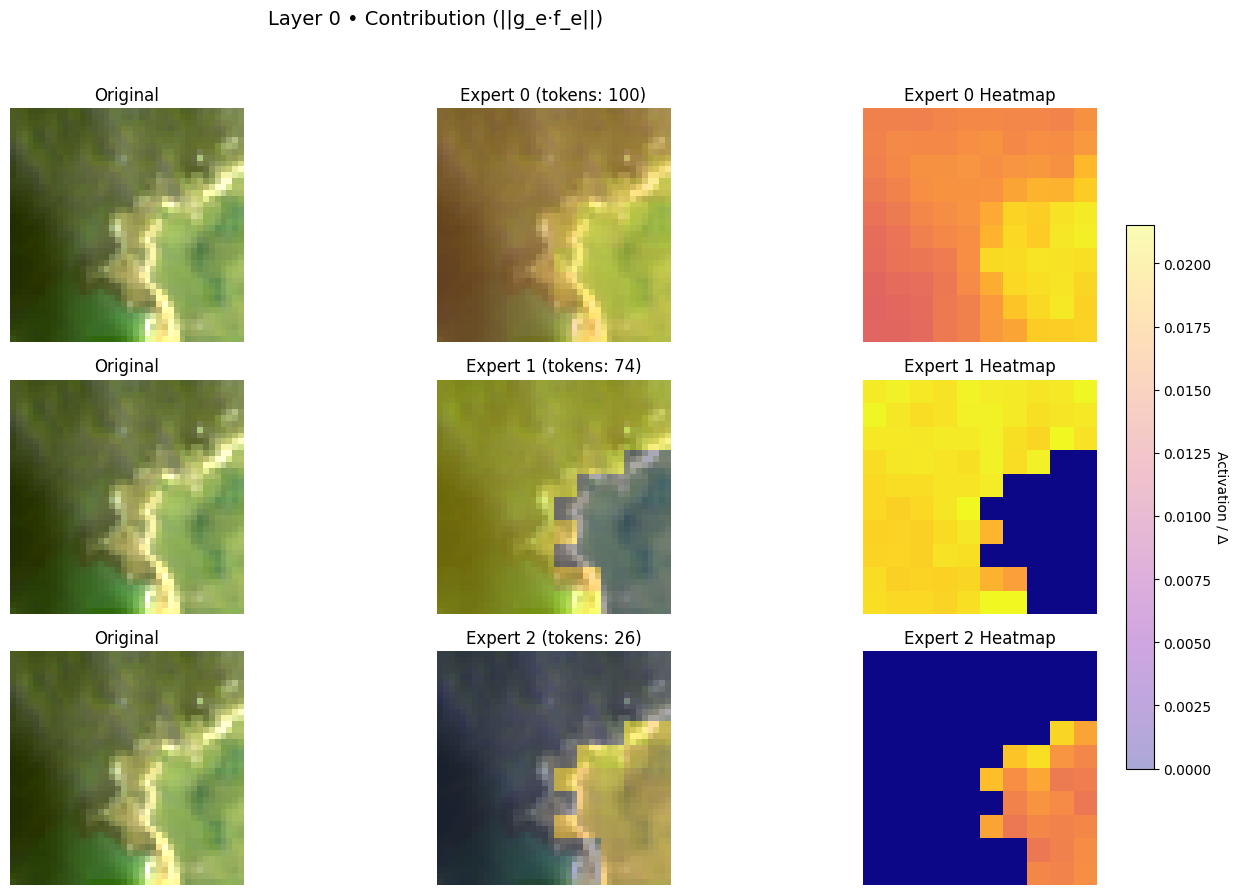} &
        \includegraphics[width=0.30\linewidth]{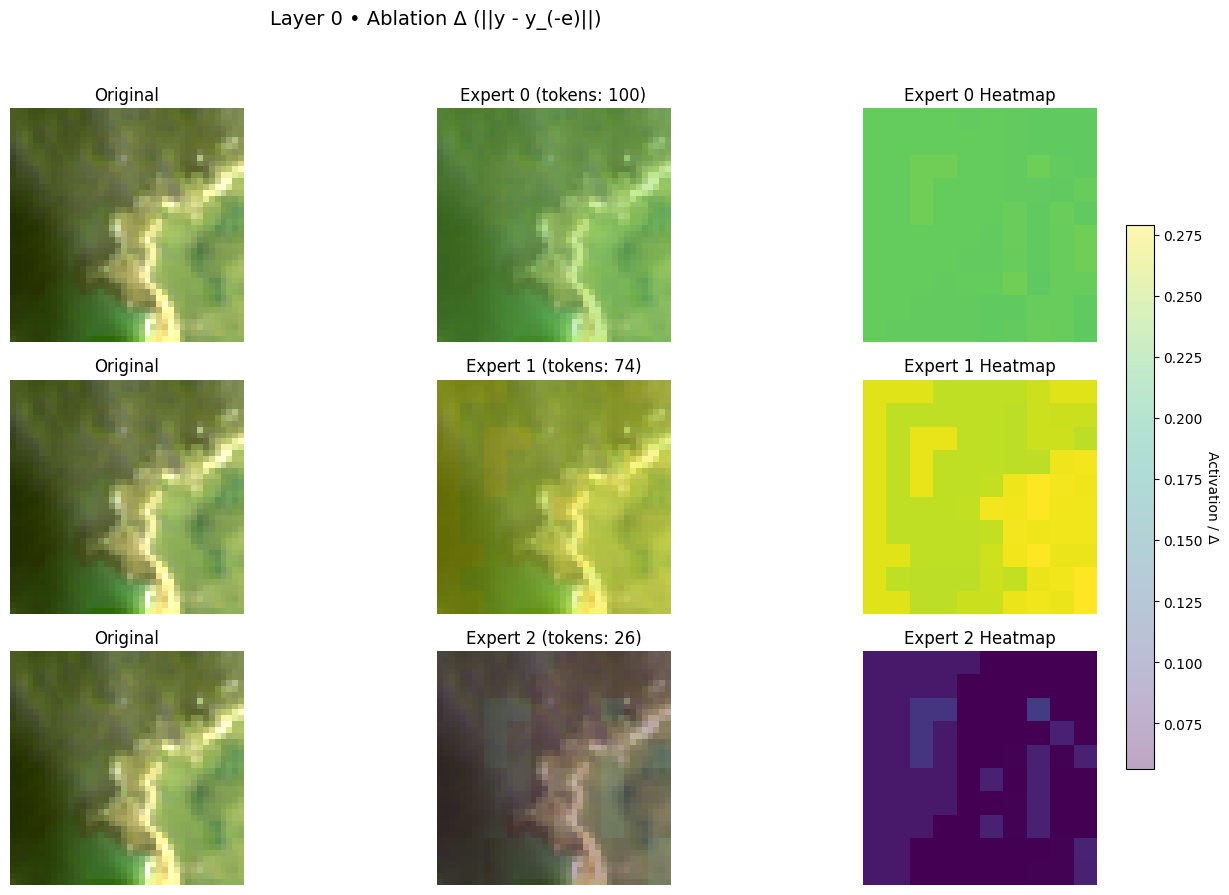} \\
    \end{tabular}
    \caption{Visualization of expert specialization at the first MoE layer across five examples. Each row corresponds to one example, showing (left) top-1 expert assignment per patch, (middle) expert contribution maps, and (right) ablation $\Delta$ maps.}
    \label{fig:moe_examples}
\end{figure}

Different examples from BEN-LS test split are used for the ablation study Fig.\ref{fig:moe_examples} depicts five of these examples and the ablation results of the first MoE layer, which consists of three experts. These results show a consistent specialization of these experts. To be more specific, Expert~$0$ is mainly routed to patches dominated by vegetation.  Expert~$1$ is consistently activated over water and shadow areas, capturing low-reflectence patterns. Finally, Expert~$2$ is often used over textured regions and contributes strongly to patches with mixed land cover. The removal of one of these experts causes localized changes in most of the cases, suggesting their complementary role. Together, these results highlight that the experts are not redundant: the same experts repeatedly specialize in semantically distinct patterns across different inputs.

\paragraph{Sparsity analysis.} 
To assess inference-time encoder sparsity, we measure the fraction of parameters activated per token (see Table~\ref{tab:ffn_sparsity_layer}). We find that each token uses about 81\% of the encoder’s parameters—covering the always-on embeddings, attention layers, and shared projections. In contrast, top-k routing activates only about 52\% of the capacity in the expert-specific feed-forward networks (FFNs) per token. This shows that sparsity is achieved where it matters most, \textit{i.e.,} the costly FFNs, thereby reducing per-token compute.

\begin{table}[ht]
\centering
\caption{Per-layer FFN (expert-unique) sizes and sparsity ratio.}
\label{tab:ffn_sparsity_layer}
\begin{tabular}{rccrc}
\toprule
\textbf{Layer} & \textbf{E} & \textbf{k} & \textbf{Unique FFN} ($\times 10^3$) & \textbf{$k/E$} \\
\midrule
0 & 3 & 2 & 62.6 & 0.67 \\
1 & 3 & 2 & 60.0 & 0.67 \\
2 & 3 & 2 & 57.9 & 0.67 \\
3 & 3 & 2 & 55.7 & 0.67 \\
4 & 3 & 2 & 53.5 & 0.67 \\
5 & 4 & 2 & 68.4 & 0.50 \\
6 & 4 & 2 & 65.5 & 0.50 \\
7 & 4 & 2 & 62.6 & 0.50 \\
8 & 4 & 2 & 59.2 & 0.50 \\
9 & 4 & 2 & 56.3 & 0.50 \\
10 & 5 & 2 & 66.7 & 0.40 \\
11 & 5 & 2 & 63.1 & 0.40 \\
12 & 5 & 2 & 59.5 & 0.40 \\
13 & 5 & 2 & 55.8 & 0.40 \\
14 & 5 & 2 & 52.2 & 0.40 \\
\midrule
\multicolumn{3}{r}{Total unique FFN} & 899.0 & -- \\
\bottomrule
\end{tabular}
\end{table}

\section{Conclusion}
We introduce a compact, metadata-aware mixture-of-experts masked autoencoder (MoE-MAE) for Earth observation with an encoder of $\sim$2.5M parameters, whose encoder has only $\sim$2.3M parameters. Using sparse expert routing and geo-temporal conditioning, the model achieves competitive transfer performance on BEN-LS and EuroSAT-LS while being orders of magnitude smaller than existing foundation models. Ablation studies show clear expert specialization, indicating that lightweight MoE designs can balance efficiency and representational capacity.
Looking ahead, we plan to scale training to the multi-sensor datasets commonly used for large EO foundation models, while preserving the lightweight MoE-MAE design. This will enable more direct comparisons to state-of-the-art systems across a broader range of downstream tasks. A natural extension is to incorporate temporal modeling, allowing compact models to capture dynamic Earth processes without sacrificing efficiency.

\bibliographystyle{unsrt}  
\bibliography{references}
\end{document}